\documentclass[times,twocolumn,final,authoryear]{elsarticle}

\usepackage{prletters}

\usepackage{float}
\usepackage{amsmath}
\usepackage{apalike}
\usepackage{graphicx}
\usepackage{caption}
\usepackage{subcaption}
\graphicspath{ {./} }

\usepackage{url}
\usepackage{xcolor}
\usepackage[acronym]{glossaries}
\usepackage{amsfonts}
\usepackage{colortbl}
\usepackage{hyperref}

\journal{Pattern Recognition Letters}

\newacronym{map}{mAP}{mean Average Precision}
\newacronym{ap}{AP}{Average Precision}
\newacronym{dpo}{DPO}{Direct Preference Optimization}
\newacronym{llm}{LLM}{Large Language Model}
\newacronym{lvlm}{LVLM}{Large Vision-Language Model}
\newacronym{ppo}{PPO}{Proximal Policy Optimization}
\newacronym{rlhf}{RLHF}{Reinforcement Learning from Human Feedback}
\newacronym{grpo}{GRPO}{Group Relative Policy Optimization}
\newacronym{cot}{CoT}{Chain-of-Thought}
\newacronym{lora}{LoRA}{Low-Rank Adaptation}
\newacronym{rl}{RL}{Reinforcement Learning}
\newacronym{kl}{KL}{Kullback-Leibler}

\begin{document}

    \thispagestyle{empty}

    \title{Entity Re-identification in Visual Storytelling via Contrastive Reinforcement Learning}

    \author[inesc,ist]{Daniel Oliveira}
    \ead{daniel.oliveira@inesc-id.pt}
    \author[inesc,ist]{David Martins de Matos}
    \ead{david.matos@inesc-id.pt}

    \address[inesc]{INESC-ID Lisboa, R. Alves Redol 9, 1000-029 Lisboa, Portugal}
    \address[ist]{Instituto Superior Técnico, Universidade de Lisboa, Av. Rovisco Pais, 1049-001 Lisboa, Portugal}

    \begin{abstract}
        Visual storytelling systems, particularly large vision-language models, struggle to maintain character and object identity across frames,
        often failing to recognize when entities in different images represent the same individuals or objects,
        leading to inconsistent references and referential hallucinations.
        This occurs because models lack explicit training on when to establish entity connections across frames.
        We propose a contrastive reinforcement learning approach that trains models to discriminate between coherent image sequences
        and stories from unrelated images.
        We extend the Story Reasoning dataset with synthetic negative examples to teach appropriate entity connection behavior.
        We employ Direct Preference Optimization with a dual-component reward function that promotes grounding and re-identification of entities
        in real stories while penalizing incorrect entity connections in synthetic contexts.
        Using this contrastive framework, we fine-tune Qwen Storyteller (based on Qwen2.5-VL 7B).
        Evaluation shows improvements in grounding \gls{map} from 0.27 to 0.31 (+14.8\%), F1 from 0.35 to 0.41 (+17.1\%).
        Pronoun grounding accuracy improved across all pronoun types except ``its'',
        and cross-frame character and object persistence increased
        across all frame counts, with entities appearing in 5 or more frames advancing from 29.3\% to 33.3\% (+13.7\%).
        Well-structured stories, containing the chain-of-thought and grounded story, increased from 79.1\% to 97.5\% (+23.3\%).
    \end{abstract}


    \maketitle

    \section{Introduction}\label{sec:introduction}
    Visual storytelling systems, while demonstrating substantial progress in generating narratives from image sequences, continue
    to struggle with maintaining consistent entity references and achieving reliable grounding of textual elements to visual counterparts~\citep{Oliveira2024StoryGF}.
    Current approaches face challenges in maintaining consistent entity identity across frames, leading to models
    that fail to properly re-identify characters and objects across temporal sequences~\citep{tacl_a_00553}.
    Even state-of-the-art \glspl{lvlm} trained on carefully curated datasets exhibit limitations in cross-frame entity re-identification,
    frequently hallucinating non-existent objects and failing to recognize when entities appearing in different frames represent the same individuals or
    objects~\citep{farquhar2024, Liu2024ASO}.
    Existing supervised approaches for visual storytelling train primarily on positive examples from coherent sequences,
    lacking negative pairs that could teach models when not to establish cross-frame entity
    connections~\citep{huang2016visual, Yu_2021_CVPR, wang2018no, Oliveira2025GroundCapAV}.
    This leads to false connections when visually similar entities appear across unrelated images.
    We propose a contrastive reinforcement learning framework using synthetic negative examples to improve cross-frame entity
    re-identification and visual grounding by promoting connections in coherent sequences while discouraging them in synthetic arrangements.

    We build upon the Story Reasoning dataset~\citep{Oliveira2025StoryReasoning}, which provides structured entity tracking and grounding
    annotations.
    The entity re-identification approach used to generate the Story Reasoning dataset relies primarily on visual similarity within cropped
    bounding boxes, without considering the whole image context.
    This can lead to incorrect connections between visually similar but contextually distinct entities.
    For instance, two cars of the same color could be misidentified across different frames based solely on visual similarity.
    This problem could be mitigated by incorporating broader contextual information, such as the relative position of objects within scenes,
    their surroundings, and other environmental cues that distinguish between similar-looking but distinct entities.
    We extend the StoryReasoning dataset with synthetic negative stories constructed by deterministically sampling images from different movies,
    creating incoherent sequences that provide negative samples, teaching models when cross-frame entity connections should not be established.
    We develop a dual-component reward function that combines re-identification accuracy and grounding quality to
    encourage appropriate entity connections in coherent narrative sequences while penalizing connections in synthetic negative stories.

    Our main contributions are:
    (1) a synthetic story generation method that creates negative examples from frames from unrelated movies;
    (2) a reward function that promotes grounding and entity re-identification in real stories while penalizing it in synthetic negative ones;
    (3) a contrastive \gls{rl} framework that uses negative examples to improve cross-frame entity re-identification and grounding.

    The remainder of this paper is organized as follows:
    Section~\ref{sec:related-work} reviews related work,
    Section~\ref{sec:methodology} describes our contrastive reinforcement learning approach,
    Section~\ref{sec:evaluation-results} presents evaluation results,
    Section~\ref{sec:limitations} discusses limitations,
    and Section~\ref{sec:conclusions-and-future-work} provides conclusions and outlines future work.

    \section{Related Work}\label{sec:related-work}
    This section reviews relevant advances in visual storytelling, contrastive learning and reinforcement
    learning for vision-language tasks.

    \subsection{Visual Storytelling and Cross-Frame Consistency}\label{subsec:visual-storytelling-consistency}
    Visual storytelling extends beyond image captioning by generating narratives that connect multiple images through temporal and causal relationships.
    Early approaches used sequential RNN architectures~\citep{huang2016visual} but struggled with character consistency and narrative coherence.
    Recent work has focused on improving narrative quality through hierarchical approaches and attention mechanisms.
    TAPM~\citep{Yu_2021_CVPR} introduced transitional adaptation for better visual-textual alignment, while CharGrid~\citep{tacl_a_00553}
    implicitly models character relationships across frames.
    TARN-VIST~\citep{chen2024tarn} employs topic-aware reinforcement learning with dual rewards to enhance narrative coherence
    by incorporating latent topic information from both visual and linguistic perspectives.
    \cite{song2024context} proposed a framework using visual prefix tuning with multimodal contrastive
    objectives to improve visual grounding and story informativeness.

    GroundCap~\citep{Oliveira2025GroundCapAV} provides 52k movie images with an ID-based grounding system that links text spans directly to visual entities
    through specialized tags, maintaining object identity across multiple references within individual images.
    Story Reasoning extends this to cross-frame consistency with 4.2k stories from movie sequences, incorporating structured
    scene analyses in the form of \gls{cot} and grounded stories.
    The \gls{cot} tracks entities through structured tabular representations,
    where each character and object is assigned a persistent identifier that remains consistent across all
    frames in which that entity appears.
    These tabular representations include bounding box coordinates for each entity instance, linking the spatial location
    of every appearance to the global entity identifier.
    Stories reference these identifiers through XML tags that include image demarcation tags
    (``\texttt{<gdi>}'' for wrapping story segments corresponding to each input image),
    entity tags (``\texttt{<gdo>}'' for characters and objects), action tags
    (``\texttt{<gda>}'' for linking actions to actors), and location tags (``\texttt{<gdl>}'' for landmarks).
    This framework creates explicit connections between narrative elements and their corresponding visual entities,
    enabling coherent storytelling and maintaining identities throughout the story.
    This entity re-identification relies on visual similarity within cropped regions, without considering
    the broader image context, which may result in incorrect connections between visually similar but contextually distinct entities.

    Contrastive learning has emerged as a powerful paradigm for vision-language understanding.
    CLIP~\citep{radford2021learning} demonstrated that simple contrastive pre-training on 400 million image-text pairs enables zero-shot
    transfer to downstream tasks.
    The method trains paired encoders to maximize similarity between matching image-text pairs while minimizing similarity for non-matching pairs.
    ALIGN~\citep{jia2021scaling} scaled contrastive learning to over one billion image-text pairs, DeCLIP~\citep{li2022supervision}
    and CLOOB~\citep{cao2022cloob} introduced improved distance metrics for handling dataset noise.
    However, these methods focus on single image-text alignment rather than sequential narrative generation,
    limiting their applicability to cross-frame consistency challenges in visual storytelling tasks 
    where maintaining entity identity across multiple frames is crucial.

    Reinforcement learning has shown promise for optimizing vision-language models beyond differentiable metrics.
    Early applications to vision-language tasks used policy gradient and actor-critic methods~\citep{rennie2017self}
    to optimize non-differentiable metrics like BLEU and CIDEr scores in image captioning.
    \gls{ppo}~\citep{schulman2017proximal} introduced clipped surrogate objectives that balance sample efficiency with training stability.
    \gls{ppo} has become central to \gls{rlhf} pipelines~\citep{ouyang2022training, christiano2017deep}, where models are fine-tuned using
    human preference feedback to improve alignment with human evaluators.
    \gls{dpo}~\citep{rafailov2023direct} emerged as a simpler alternative to \gls{rlhf}, directly optimizing policies using preference pairs
    through a classification objective offering improved stability over on-policy methods like \gls{ppo} through its off-policy formulation.

    \section{Methodology}\label{sec:methodology}
    Our contrastive reinforcement learning approach improves cross-frame entity re-identification and grounding capabilities
    by training models to establish entity connections only on real stories.
    Our method uses differential rewards to encourage proper entity tracking in coherent sequences while discouraging spurious connections in incoherent arrangements.
    This section details our synthetic story generation methodology, reward function design, and \gls{dpo} training framework.

    \subsection{Synthetic Story Generation}\label{subsec:synthetic-story-generation}
    We extend the Story Reasoning dataset~\citep{Oliveira2025StoryReasoning} by algorithmically generating synthetic stories that serve
    as negative examples for contrastive training with a 2:1 ratio of real to synthetic stories.
    For each synthetic story, we deterministically select between 5 and 15 images from different real stories
    using a sampling algorithm designed to be deterministic for reproducibility while ensuring visual incoherence between selected images.
    Given a synthetic story index $s$ and desired frame count $n$, we select images using $\text{story\_idx}_i = (s \times 17 + i \times 31) \bmod N$
    and $\text{img\_idx}_i = (s + i \times 7) \bmod |\mathcal{I}_{\text{story\_idx}_i}|$, where $\text{story\_idx}_i$ identifies the source
    story from which to sample the $i$-th frame, $\text{img\_idx}_i$ specifies
    which image within that source story to select, $N$ is the total number of real stories, $\mathcal{I}_j$ represents the
    image set for story $j$, and $i \in [0, n-1]$ iterates through frame positions of the synthetic story.

    The algorithm intentionally selects images from stories that are far apart in the dataset ordering,
    minimizing the likelihood of the selected images belonging to the same movie and thus ensuring visual incoherence.
    This synthetic dataset construction doubles the original dataset size, creating 4,178 synthetic stories alongside the 4,178
    real stories.
    This provides equal exposure to positive and negative examples during training.

    \subsection{Reward Function}\label{subsec:reward-function}
    We design a dual-component reward function that promotes desirable behaviors for real stories while penalizing the same behaviors in synthetic stories.
    Following the approach introduced in DeepSeek-R1~\citep{deepseek2025r1}, we employ rule-based rewards to avoid common reward hacking
    issues common with neural reward models.
    Our reward function combines entity re-identification ($R_{\text{reid}}$) and grounding ($R_{\text{ground}}$) with structural validation
    to ensure generated outputs conform to the expected format.
    The reward function first validates the structural integrity of both the \gls{cot} and the generated story against the input images.
    If the generated content violates structural constraints or contains formatting errors, the function returns a penalty score of -1.0.
    For structurally valid outputs, the function computes the weighted combination of the two reward components as shown in Eq.~\ref{eq:reward_function}.

    \begin{equation}
        R(c, s, \mathcal{I}, r) = \begin{cases}
                                      0.5 \times R_{\text{reid}}(c, r) + 0.5 \times R_{\text{ground}}(s) & \text{if valid} \\
                                      -1.0 & \text{if invalid}
        \end{cases}
        \label{eq:reward_function}
    \end{equation}

    In Eq.~\ref{eq:reward_function} $c$ represents the \gls{cot}, $s$ is the generated story,
    $\mathcal{I}$ denotes the input images, and $r$ indicates whether the story is real or synthetic.

    \subsubsection{Structure Validation}
    We implement structure validation to ensure generated outputs maintain the required format and consistency.
    Our validation process consists of two main components:

    \textbf{\gls{cot} Validation:} We validate the structured analysis by checking that:
    (1) each input image has a corresponding analysis section;
    (2) character identifiers follow the correct format, such as ``\texttt{char1}'' and ``\texttt{char2}'';
    (3) object identifiers use proper prefixes, ``\texttt{obj}'' for objects, ``\texttt{lm}'' for landmarks, and ``\texttt{bg}'' for background elements;
    (4) bounding box coordinates fall within image boundaries;
    (5) all five narrative phases are present (Introduction, Development, Conflict, Turning Point, Conclusion); and
    (6) character, object, and setting metadata tables maintain proper structure with required columns.

    \textbf{Story Validation:} We validate the generated narrative by ensuring:
    (1) the number of ``\texttt{<gdi image*>}'' tags matches the number of input images; and
    (2) all entity IDs referenced in grounding tags (``\texttt{<gdo>}'', ``\texttt{<gda>}'', ``\texttt{<gdl>}'')
    appear in the corresponding \gls{cot} table entries, ensuring consistency between the \gls{cot} and the generated story.

    Following the same approach as DeepSeek-R1~\citep{deepseek2025r1}, responses that fail any validation check receive a penalty of -1.0.

    \subsubsection{Entity Re-identification Reward}
    The entity re-identification component measures cross-frame consistency by tracking character and object persistence across the sequence
    as shown in Eq.~\ref{eq:reid_reward}.

    \begin{equation}
        R_{\text{reid}}(c, r) = \begin{cases}
                                    \alpha \times R_{\text{char}} + \beta \times R_{\text{obj}} & \text{if } r = \text{True} \\
                                    1.0 - (\alpha \times R_{\text{char}} + \beta \times R_{\text{obj}}) & \text{if } r = \text{False}
        \end{cases}
        \label{eq:reid_reward}
    \end{equation}

    In Eq.~\ref{eq:reid_reward} $c$ represents the \gls{cot}, $r$ indicates whether the story is real (true) or synthetic (false),
    $\alpha$ and $\beta$ are weighting parameters that control the relative importance of character versus object re-identification.
    We set $\alpha = 0.6$ and $\beta = 0.4$ to prioritize character re-identification, as characters typically drive narrative progression.
    These values can be adjusted as needed to balance the focus between character and object re-identification.
    The character re-identification score $R_{\text{char}}$ and object re-identification score $R_{\text{obj}}$ are computed as shown
    in Eq.~\ref{eq:char_reward} and Eq.~\ref{eq:obj_reward}.

    \begin{equation}
        R_{\text{char}} = \min\left(1.0, \frac{\sum_{c_i \in \mathcal{C}} |\mathcal{F}_{c_i}|}{|\mathcal{C}| \times |\mathcal{I}|}\right)
        \label{eq:char_reward}
    \end{equation}

    \begin{equation}
        R_{\text{obj}} = \min\left(1.0, \frac{\sum_{o_j \in \mathcal{O}} |\mathcal{F}_{o_j}|}{|\mathcal{O}| \times |\mathcal{I}|}\right)
        \label{eq:obj_reward}
    \end{equation}

    $\mathcal{C}$ and $\mathcal{O}$ represent the sets of detected characters and objects, $\mathcal{F}_{c_i}$ and $\mathcal{F}_{o_j}$ denote the frame sets where character $c_i$ and object $o_j$ appear, and $|\mathcal{I}|$ is the total number of frames.

    This formulation rewards models for re-identifying entities across frames in authentic stories while penalizing re-identification
    of entities in synthetic stories, encouraging the model to develop robust discrimination capabilities for when cross-frame entity tracking is appropriate.

    \subsubsection{Pronoun Grounding Reward}
    The pronoun grounding component evaluates whether the model appropriately grounds subsequent references to entities,
    rewarding cases where pronouns and proper nouns maintain explicit connections to their corresponding visual entities.
    The reward shown in Eq.~\ref{eq:grounding_reward}.

    \begin{equation}
        R_{\text{grounding}}(s) = \gamma \times \frac{G_{\text{char}} + P_{\text{char}}}{T_{\text{char}}} + \delta \times \frac{G_{\text{obj}} + P_{\text{obj}}}{T_{\text{obj}}}
        \label{eq:grounding_reward}
    \end{equation}

    Where $G_{\text{char}}$ and $G_{\text{obj}}$ represent grounded pronouns for characters and objects, $P_{\text{char}}$ and $P_{\text{obj}}$
    denote grounded proper nouns, $T_{\text{char}}$ and $T_{\text{obj}}$ indicate total pronouns and proper nouns in the story,
    and $\gamma$ and $\delta$ are weighting parameters controlling the relative importance of character versus object grounding.
    We set $\gamma = 0.5$ and $\delta = 0.5$ to equally weight character and object grounding, as we do not assume a preference for either entity type.
    This component encourages the model to maintain traceable references throughout the narrative,
    ensuring that pronouns like ``he'', ``she'', or ``they'' can be linked back to specific visual entities rather than creating
    ambiguous references.
    We extract grounded references using regular expressions to identify entity tags, then employ spaCy for part-of-speech analysis
    to classify the content within those tags as pronouns or proper nouns.
    The grounding reward encourages entity-text alignment regardless of story authenticity.

    \subsection{Direct Preference Optimization Training}\label{subsec:dpo-training}
    \glsreset{dpo}
    \gls{dpo}~\citep{rafailov2023direct} further fine-tunes Qwen Storyteller using preference pairs generated offline from the contrastive reward function in Eq.~\ref{eq:reward_function}.
    Qwen Storyteller is a \gls{lora}~\citep{hu2022lora} rank 2048 fine-tuned version of Qwen2.5-VL 7B that was initially trained on the Story Reasoning dataset through supervised fine-tuning.
    \gls{dpo} directly optimizes the policy using preference data without requiring explicit reward model training, offering improved stability over \gls{rlhf} that rely on the \gls{ppo},
    loss function is shown in Eq.~\ref{eq:dpo_objective}.

    \begin{equation}
        L_{\text{DPO}}(\pi_\theta) = -\mathbb{E}_{(x,y_w,y_l) \sim \mathcal{D}} \left[ \log \sigma \left( \beta \log \frac{\pi_\theta(y_w|x)}{\pi_{\text{ref}}(y_w|x)} - \beta \log \frac{\pi_\theta(y_l|x)}{\pi_{\text{ref}}(y_l|x)} \right) \right]
        \label{eq:dpo_objective}
    \end{equation}

    In Eq.~\ref{eq:dpo_objective} $x$ represents the input image sequence, $y_w$ and $y_l$ are the chosen and rejected responses respectively,
    $\pi_\theta$ is the policy being trained, $\pi_{\text{ref}}$ is the reference policy (initial supervised fine-tuned model),
    $\beta$ is the temperature parameter controlling the \gls{kl} constraint strength, $\sigma$ is the sigmoid function $\sigma(z) = \frac{1}{1 + e^{-z}}$, and
    $\mathcal{D}$ is the preference dataset containing triplets of input sequences and preference pairs.

    Our approach generates preference pairs offline by sampling multiple responses for each image sequence and ranking them using the reward function from Eq.~\ref{eq:reward_function}.
    For each story, one preference pair is generated where the chosen response is guaranteed to have a reward at least 0.05 higher than the rejected response.
    The implicit \gls{kl} regularization in the \gls{dpo} objective ensures that the fine-tuned model does not deviate excessively
    from the reference model, maintaining its storytelling capabilities while learning improved entity re-identification behavior.
    When processing real stories, preference pairs favor responses with higher entity re-identification and grounding scores.
    When processing synthetic stories, pairs favor responses with lower re-identification scores, teaching the model to avoid inappropriate cross-frame connections.

    Two training experiments are conducted to evaluate the effectiveness of this approach.
    The first experiment employs \gls{lora} with rank 2048 and alpha scaling factor 4096 for parameter-efficient training,
    targeting self-attention layers in the language components.
    The second experiment uses full fine-tuning to assess the impact of training all model parameters.
    Both experiments employee the temperature parameter $\beta = 0.1$, sigmoid loss function,
    AdamW~\citep{loshchilov2018decoupled} optimizer with learning rate $5 \times 10^{-6}$, batch size 8, and 3 training epochs.

    \section{Evaluation Results}\label{sec:evaluation-results}
    We evaluate the contrastive reinforcement learning approach using automatic metrics that assess grounding effectiveness,
    entity re-identification performance, and linguistic quality.
    Evaluation compares both \gls{lora} and full fine-tuning configurations against the baseline Qwen Storyteller model
    \footnote{The trained models and dataset are available at: \url{https://huggingface.co/datasets/daniel3303/StoryReasoningAdversarialDPO}
    and \url{https://huggingface.co/daniel3303/QwenStoryteller2}}.

    \subsection{Automatic Metrics}\label{subsec:automatic-metrics}
    We evaluate grounding effectiveness using precision ($P = \frac{TP}{TP+FP}$), recall ($R = \frac{TP}{TP+FN}$),
    and F1 score ($F1 = 2 \cdot \frac{P \cdot R}{P+R}$) for entity references in generated stories, 
    where TP/FP/TN/FN denote true/false positive/negative predictions.
    We use an adaptation of \gls{map} described in~\citep{Oliveira2025StoryReasoning}, calculating Average Precision for each
    story using 11-point interpolation, then averaging across all stories.

    We measure entity persistence by tracking characters and objects that appear across multiple frames, analyzing re-identification patterns
    for both authentic and synthetic stories.
    We also report standard language metrics (METEOR~\citep{banerjee2005meteor}, ROUGE-L~\citep{lin-2004-rouge}, BLEU-4~\citep{bleu2002bleu}) to assess narrative quality changes.

    Table~\ref{tab:automatic-metrics} presents results comparing our contrastive reinforcement learning models against the baseline Qwen Storyteller.

    \begin{table*}[ht]
        \centering
        \caption{Automatic evaluation results comparing contrastive reinforcement learning approaches with baseline Qwen Storyteller. Precision and Recall reported for character (Char), object (Obj), and combined entity (Total) references. \cellcolor{green!20} Best values and \cellcolor{red!20} worst values are highlighted.}
        \label{tab:automatic-metrics}
        \begin{tabular}{l|cccc|ccc|c|ccc}
            \hline
            & \multicolumn{4}{c|}{Precision} & \multicolumn{3}{c|}{Recall} & F1 & \multicolumn{3}{c}{Language} \\
            Model                                    & Char                     & Obj                      & Total                    & \gls{map}                & Char                     & Obj                      & Total                    & Total                    & M                        & R                        & B-4                       \\
            \hline
            Baseline (Qwen Storyteller)              & 0.83                     & \cellcolor{green!20}0.46 & 0.57                     & \cellcolor{red!20}0.27   & \cellcolor{red!20}0.62   & \cellcolor{red!20}0.25   & \cellcolor{red!20}0.40   & \cellcolor{red!20}0.35   & \cellcolor{red!20}0.14   & 0.16                     & 0.054                           \\
            Contrastive \gls{rl} (\gls{lora} R=512)  & \cellcolor{green!20}0.84 & 0.45                     & \cellcolor{green!20}0.57 & 0.27                     & 0.64                     & 0.25                     & 0.41                     & 0.36                     & 0.14                     & \cellcolor{red!20}0.16   & \cellcolor{red!20}0.049         \\
            Contrastive \gls{rl} (\gls{lora} R=1024) & 0.82                     & 0.38                     & 0.52                     & 0.30                     & 0.71                     & 0.28                     & 0.45                     & 0.39                     & 0.15                     & 0.17                     & 0.053                           \\
            Contrastive \gls{rl} (\gls{lora} R=2048) & \cellcolor{red!20}0.78   & \cellcolor{red!20}0.29   & \cellcolor{red!20}0.45   & \cellcolor{green!20}0.31 & \cellcolor{green!20}0.77 & \cellcolor{green!20}0.28 & \cellcolor{green!20}0.48 & \cellcolor{green!20}0.41 & \cellcolor{green!20}0.17 & \cellcolor{green!20}0.18 & \cellcolor{green!20}0.057   \\
            \hline
        \end{tabular}
    \end{table*}

    The experimental evaluation demonstrates improvements when comparing the baseline with \gls{lora} rank 2028. 
    \gls{map} increased from 0.27 to 0.31 (+14.8\%),
    precision decreased but recall and F1 score improved from 0.40 to 0.48 (+20.0\%) and 0.35 to 0.41 (+17.1\%). 
    All the language metrics improved as well, with METEOR increasing from 0.14 to 0.17 (+21.4\%), ROUGE-L from 0.16 to 0.18 (+12.5\%), and BLEU-4 from 0.054 to 0.057 (+5.6\%).

    \subsection{Entity Re-identification Analysis}\label{subsec:entity-reidentification-analysis}
\begin{figure*}[ht]
        \centering
        \begin{subfigure}[b]{0.48\textwidth}
            \centering
            \includegraphics[width=\textwidth]{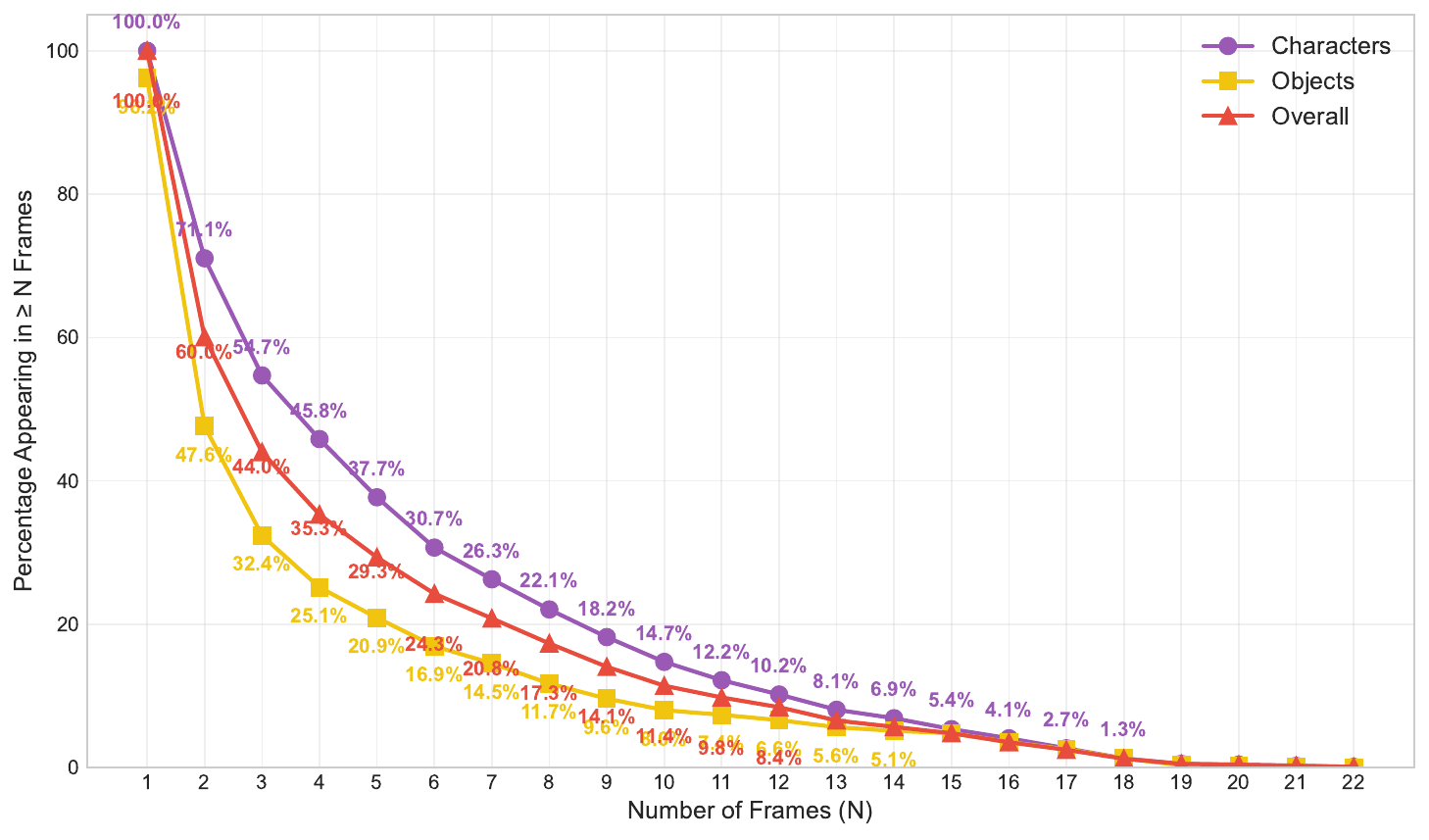}
            \caption{Baseline - Cross-frame entity persistence}
            \label{fig:baseline-cross-frame-consistency}
        \end{subfigure}
        \hfill
        \begin{subfigure}[b]{0.48\textwidth}
            \centering
            \includegraphics[width=\textwidth]{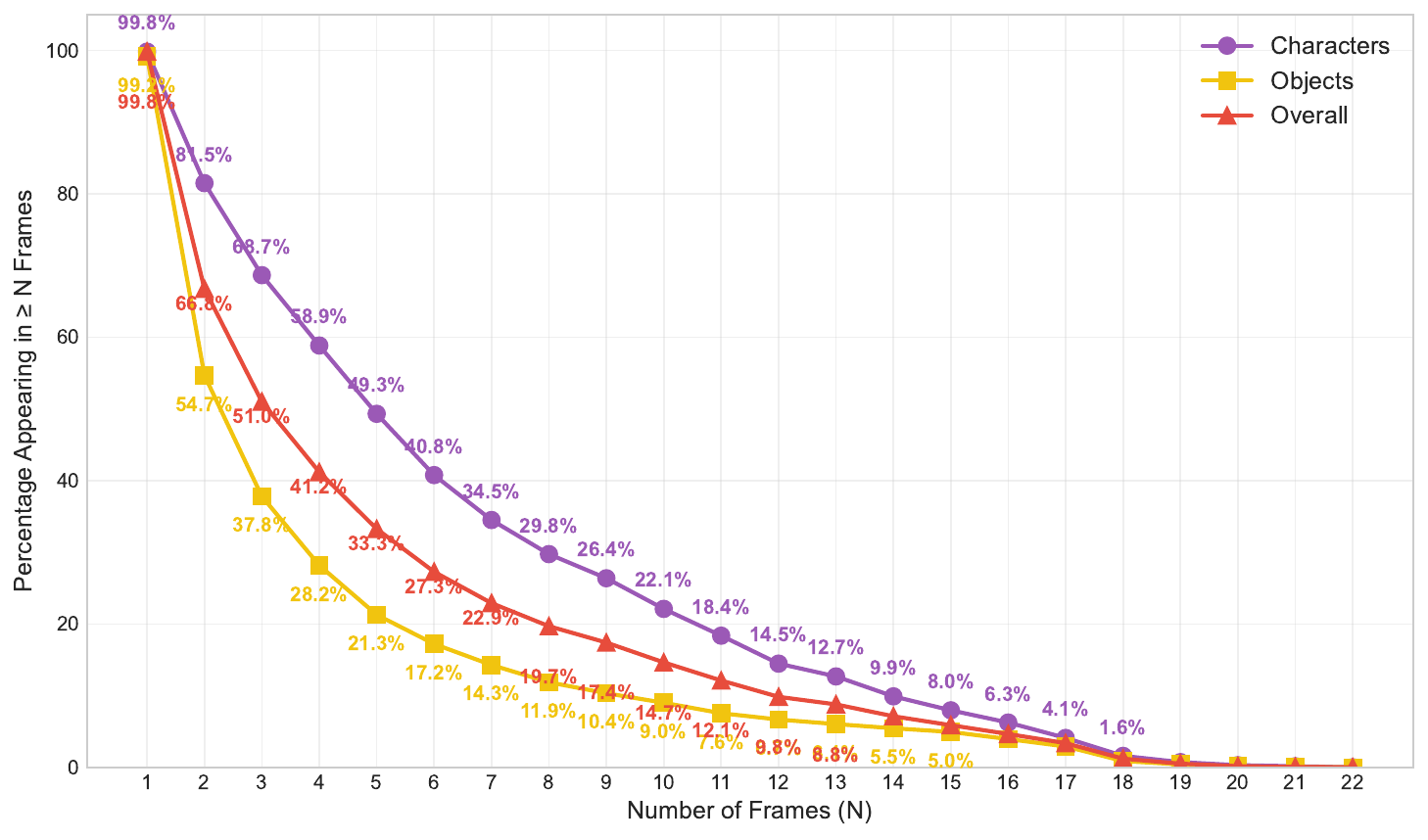}
            \caption{Contrastive \gls{rl} (LoRA R=2048) - Cross-frame entity persistence}
            \label{fig:finetuned-cross-frame-consistency}
        \end{subfigure}
        \caption{Cross-frame entity persistence comparison between baseline Qwen Storyteller model (left) and contrastive \gls{rl} model with \gls{lora} fine-tuning R=2048 (right), showing what percentage of all entities from all stories appear in N or more frames, demonstrating improved entity re-identification across all frame counts.}
        \label{fig:entity-persistence-comparison}
    \end{figure*}

    Fig.~\ref{fig:entity-persistence-comparison} illustrates entity persistence patterns across different frame counts for the
    contrastive reinforcement learning model compared to the baseline Qwen Storyteller.
    The figure shows the percentage of all entities across all stories that appear in at least N frames: purple lines represent characters, 
    yellow lines represent objects, and red lines represent the combined total.
    The results show that the contrastive \gls{rl} model maintains 49.3\% of characters and 21.3\% of objects appearing in 5 or more frames compared
    to 37.7\% and 20.9\% for the baseline.
    
    \begin{figure*}[ht]
        \centering
        \begin{subfigure}[b]{0.48\textwidth}
            \centering
            \includegraphics[width=\textwidth]{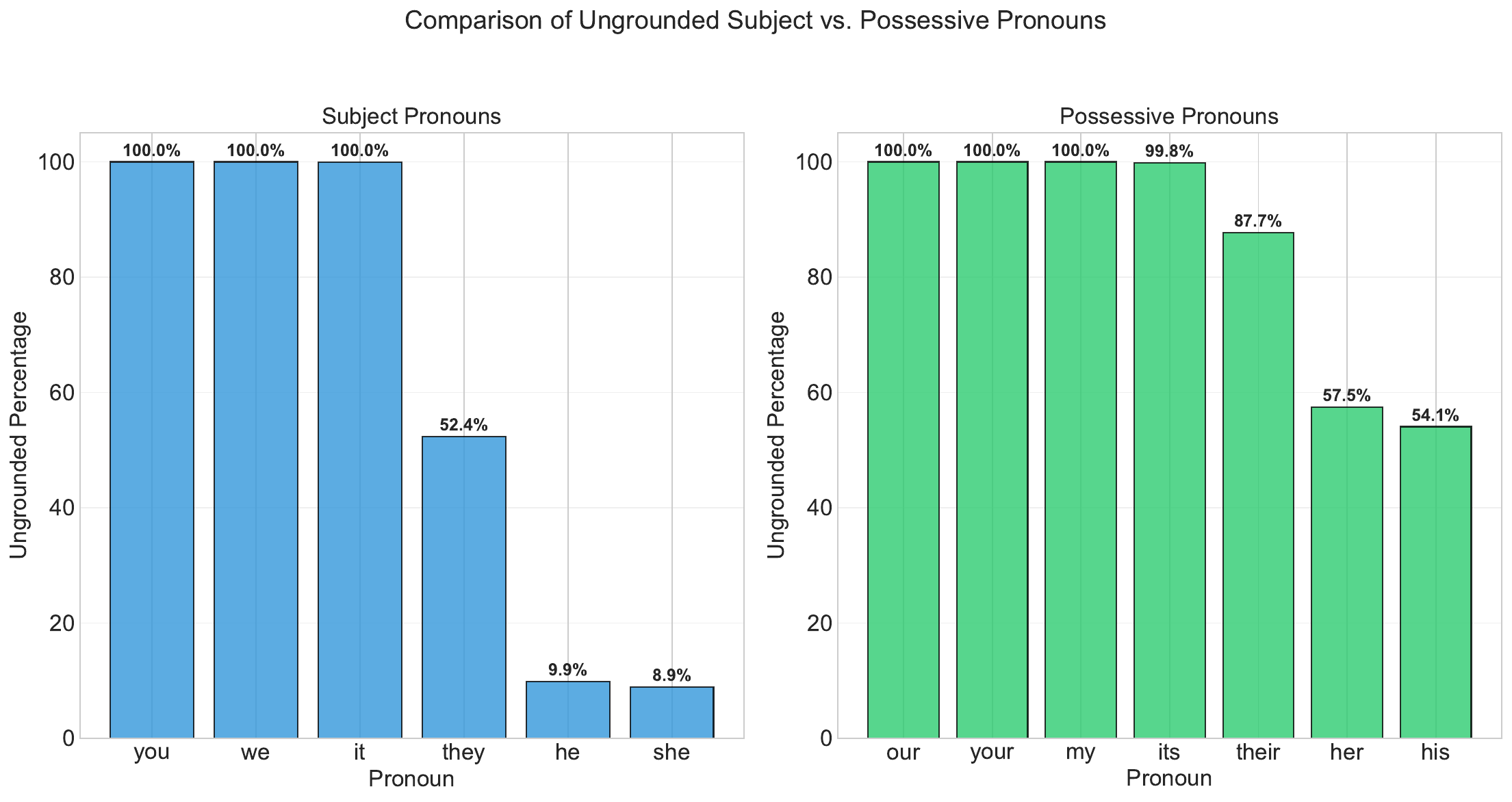}
            \caption{Baseline - Percentage of ungrounded pronouns (less is better)}
            \label{fig:baseline-subject-vs-possessive-pronouns}
        \end{subfigure}
        \hfill
        \begin{subfigure}[b]{0.48\textwidth}
            \centering
            \includegraphics[width=\textwidth]{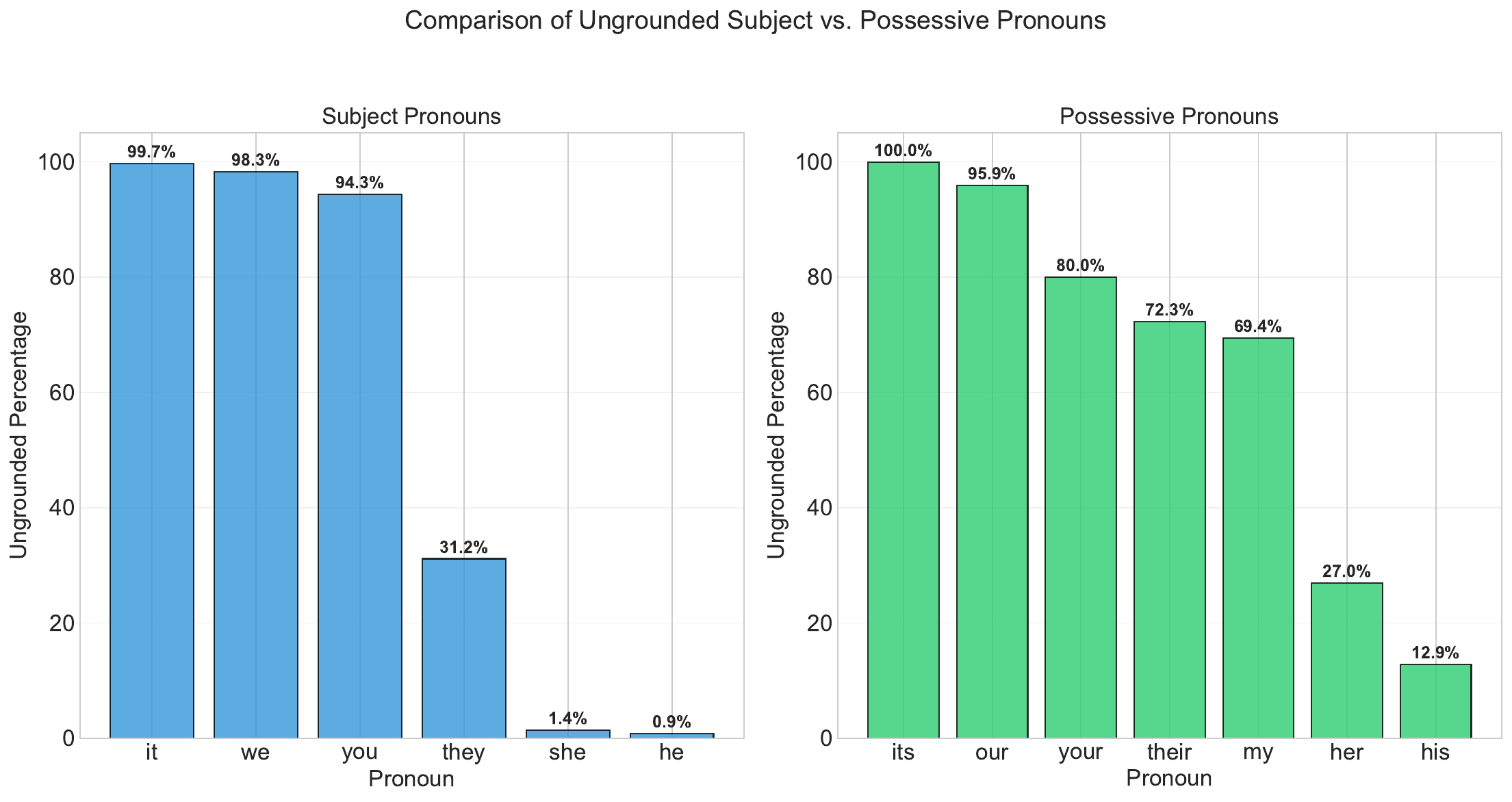}
            \caption{Contrastive \gls{rl} (LoRA R=2048) - Percentage of ungrounded pronouns (less is better)}
            \label{fig:finetuned-subject-vs-possessive-pronouns}
        \end{subfigure}
        \caption{Pronoun grounding performance comparison between baseline Qwen Storyteller model (left) and contrastive \gls{rl} model with \gls{lora} fine-tuning R=2048 (right), showing reduced percentage of ungrounded pronouns.}
        \label{fig:pronoun-grounding-comparison}
    \end{figure*}

    \subsection{Pronoun Grounding Analysis}\label{subsec:pronoun-grounding-analysis}
    We analyze pronoun grounding performance to examine how contrastive reinforcement learning improves the alignment between
    pronouns and their visual referents.
    Fig.~\ref{fig:pronoun-grounding-comparison} compares the percentage of ungrounded pronouns across different pronoun types
    for the contrastive \gls{rl} model and the baseline Qwen Storyteller.
    The baseline achieves 47.6\%, 90.1\%, and 91.1\% grounding accuracy for gender-specific pronouns ``they'', ``he'', and ``she'' respectively,
    and 12.3\%, 45.9\%, and 42.5\% for the possessive pronouns ``their'', ``his'', and ``her''.
    The contrastive \gls{rl} approaches demonstrate improvements showing 68.8\%, 99.1\%, and 98.6\% grounding accuracy for
    gender-specific ``they'', ``he'', and ``she'' respectively, and 27.7\%, 87.1\%, and 73.0\% for possessive pronouns ``their'', ``his'', and ``her''.

    These improvements show that contrastive training enhances the model's ability to maintain
    consistent pronoun-entity mappings across image sequences.
    Gender-specific pronouns (he/she, his/her) show the most relative gains, with 90.9\%/84.3\% and 76.3\%/43.0\% improvement over the baseline.
    Plural pronouns (they/their) achieve 21.2\% and 18.3\% improvement over the baseline.
    Pronouns such as ``I'', ``We'', ``You'', ``My'', ``Our'', and ``Your'' show the least improvement as they typically
    appear in character dialogues.
    These results suggest that the dual-component reward function encourages explicit grounding of pronouns to their visual
    counterparts, reducing ambiguous references that could lead to narrative inconsistencies.

    \subsection{Reward Component Analysis}\label{subsec:reward-component-analysis}
    Table~\ref{tab:reward-components} shows how each reward component drives model behavior across story types and training configurations.
    The re-identification component ($R_{\text{reid}}$) achieves the desired discrimination on real stories with higher scores
    on the Rank 2048 model (0.39) when compared with the baseline (0.34).
    On negative stories, the Rank 2048 model achieves a lower score (0.67) than the baseline (0.72), this indicates that
    the model would benefit from more training on negative stories.
    The grounding component ($R_{\text{ground}}$) improves across both story types, achieving the desired outcome.

    \begin{table*}[!h]
        \centering
        \caption{Reward component breakdown across models and story types.}
        \label{tab:reward-components}
        \begin{tabular}{l|cc|cc|cc}
            \hline
            \multirow{2}{*}{Model} & \multicolumn{2}{c|}{Real Stories} & \multicolumn{2}{c|}{Synthetic Stories} & \multicolumn{2}{c}{Overall Reward} \\
            & $R_{\text{reid}}$        & $R_{\text{ground}}$      & $R_{\text{reid}}$        & $R_{\text{ground}}$      & Real & Synthetic \\
            \hline
            Baseline (Qwen Storyteller)              & 0.34                     & \cellcolor{red!20}0.15   & \cellcolor{green!20}0.72   & \cellcolor{red!20}0.13   & \cellcolor{red!20}0.26   & 0.49    \\
            Contrastive \gls{rl} (\gls{lora} R=512)  & \cellcolor{red!20}0.32   & 0.19                     & 0.73                     & 0.17                     & 0.27                     & \cellcolor{green!20}0.51    \\
            Contrastive \gls{rl} (\gls{lora} R=1024) & 0.32                     & 0.20                     & 0.71                     & 0.18                     & 0.27                     & 0.50                     \\
            Contrastive \gls{rl} (\gls{lora} R=2048) & \cellcolor{green!20}0.39 & \cellcolor{green!20}0.22 & \cellcolor{red!20}0.67 & \cellcolor{green!20}0.20 & \cellcolor{green!20}0.32 & \cellcolor{red!20}0.48    \\
            \hline
        \end{tabular}
    \end{table*}

    \section{Limitations}\label{sec:limitations}
    Several limitations warrant consideration for future work.
    The movie-derived dataset introduces cinematic biases that may limit generalization to personal photos, surveillance footage,
    or user-generated content where visual coherence may be less pronounced.
    Despite improved entity re-identification performance, we do not validate whether the underlying bounding boxes accurately
    correspond to the referenced objects, potentially allowing cases where bounding boxes only partially cover objects or,
    in extreme cases, reference locations where the intended object is not present.
    The 2:1 ratio of real to synthetic stories may not represent the optimal balance for all training scenarios and could be adjusted based on model performance.
    Finally, our conclusions are limited to the 7B parameter QwenStoryteller model, and effectiveness may vary across different architectures,
    scales, or base model capabilities.

    \section{Conclusion and Future Work}\label{sec:conclusions-and-future-work}
    We introduce a contrastive reinforcement learning framework that addresses entity re-identification and grounding challenges in visual storytelling.
    By extending the Story Reasoning dataset with synthetic stories and employing a dual-component reward function,
    our approach teaches models to maximize entity connections in coherent sequences while discouraging them in incoherent arrangements.
    The contrastive framework effectively teaches models when not to establish cross-frame connections, leading to more reliable narrative
    generation.
    Our work establishes contrastive reinforcement learning as a viable approach for improving visual storytelling models,
    providing a practical framework and evidence for the benefits of explicitly training models on positive and negative examples.
    Future work could explore alternative synthetic story generation strategies, adaptive reward weighting mechanisms,
    and extension to other vision-language tasks such as video captioning and visual question answering.

    \section*{Acknowledgments}\label{sec:acknowledgments}
    Daniel Oliveira is supported by a scholarship granted by Fundação para a Ciência e Tecnologia (FCT), with reference 2021.06750.BD. Additionally, this work was supported by Portuguese national funds through FCT, with reference UIDB/50021/2020.

    \bibliographystyle{apalike}
    \bibliography{bibliography}

\end{document}